\pdfoutput=1

\documentclass[11pt]{article}

\usepackage[]{acl}

\usepackage{times}
\usepackage{latexsym}

\usepackage[T1]{fontenc}

\usepackage[utf8]{inputenc}

\usepackage{microtype}

\usepackage{hyperref}
\usepackage{amsmath} 
\usepackage{amssymb} 
\usepackage{graphicx} 
 \usepackage{xcolor}
\usepackage{threeparttable}
\usepackage{pifont}
\usepackage{multirow}
\usepackage{subfigure}
\usepackage{graphicx}
\usepackage{booktabs}
\usepackage{enumitem}
\usepackage{algorithm}
\usepackage{algpseudocode}
\usepackage{amsmath}
\usepackage{graphicx} 
\usepackage{mdframed}
\usepackage{relsize}

\usepackage{inconsolata}
\NewDocumentCommand{\bai}{ mO{} }{\textcolor{red}
{\textsuperscript{\textit{bai}}\textsf{\textbf{\small[#1]}}}}
\NewDocumentCommand{\andong}{ mO{} }{\textcolor{blue}
{\textsuperscript{\textit{andong}}\textsf{\textbf{\small[#1]}}}}
\title{DUAL-REFLECT: Enhancing Large Language Models for Reflective Translation through Dual Learning Feedback Mechanisms}

\author{
Andong Chen\thanks{Andong Chen and Lianzhang Lou contributed equally and are co-first authors. Work was done when Andong Chen was at Pengcheng Laboratory.}$^{\clubsuit}$\hspace{0.5mm}, 
 Lianzhang Lou\textsmaller{*}$^{\clubsuit}$\hspace{0.5mm}, 
\textbf{Kehai Chen}\thanks{~~Corresponding author.}$^{\spadesuit}$\hspace{0.5mm}, 
 \textbf{Xuefeng Bai}$^{\spadesuit}$\hspace{0.5mm}, 
 \textbf{Yang Xiang}$^{\clubsuit}$\hspace{0.5mm}, \\
 \textbf{Muyun Yang}$^{\spadesuit}$\hspace{0.5mm},
 \textbf{Tiejun Zhao}$^{\spadesuit}$\hspace{0.5mm}, 
 \textbf{Min Zhang}$^{\spadesuit}$\hspace{0.2mm}\hspace{1.5mm} \\
 $^\spadesuit$ School of Computer Science and Technology, Harbin Institute of Technology, China\\
$^\clubsuit$ Pengcheng Laboratory, Shenzhen, China \\
  ands691119@gmail.com,  \{loulzh, xiangy\}@pcl.ac.cn  \\
  \{chenkehai, baixuefeng, yangmuyun, tjzhao, zhangmin2021\}@hit.edu.cn, 
}

\begin{document}
\maketitle
\begin{abstract}
Recently, large language models (LLMs) enhanced by self-reflection have achieved promising performance on machine translation. The key idea is guiding LLMs to generate translation with human-like feedback.
However, existing self-reflection methods lack effective feedback information, limiting the translation performance.
To address this, we introduce a DUAL-REFLECT framework, leveraging the dual learning of translation tasks to provide effective feedback, thereby enhancing the models' self-reflective abilities and improving translation performance. The application of this method across various translation tasks has proven its effectiveness in improving translation accuracy and eliminating ambiguities, especially in translation tasks with low-resource language pairs\footnote{ Our code is available at \url{https://github.com/loulianzhang/Dual-Reflect}.}. 
\end{abstract}

\section{Introduction}
Large language models (LLMs) have recently demonstrated remarkable abilities across a variety of tasks ~\cite{bubeck2023sparks,xu2023large,zhao2023survey}. 
Notably, in the field of machine translation, LLMs have improved translation quality by adopting human-like methods of self-reflection ~\cite{shinn2023reflexion,liang2023encouraging}.
The self-reflection process primarily relies on using LLMs to iteratively refine initial drafts through feedback loops, a method that has been widely researched and explored ~\cite{shinn2023reflexion,DBLP:conf/uist/ParkOCMLB23, scheurer2022training,le2022coderl,welleck2022generating, amabile1983theoretical, flower1981cognitive,chen2023equals,simon1962architecture,chen-etal-2023-improving-low,sun2021joint}. 
The lack of effective feedback limits the self-reflective capacity of Large Language Models (LLMs), thereby affecting their continuous improvement in translation ~\cite{tyen2023llms,liang2023encouraging,lou-etal-2023-cceval}. 


\begin{figure}[!t]
\includegraphics[scale=0.29]{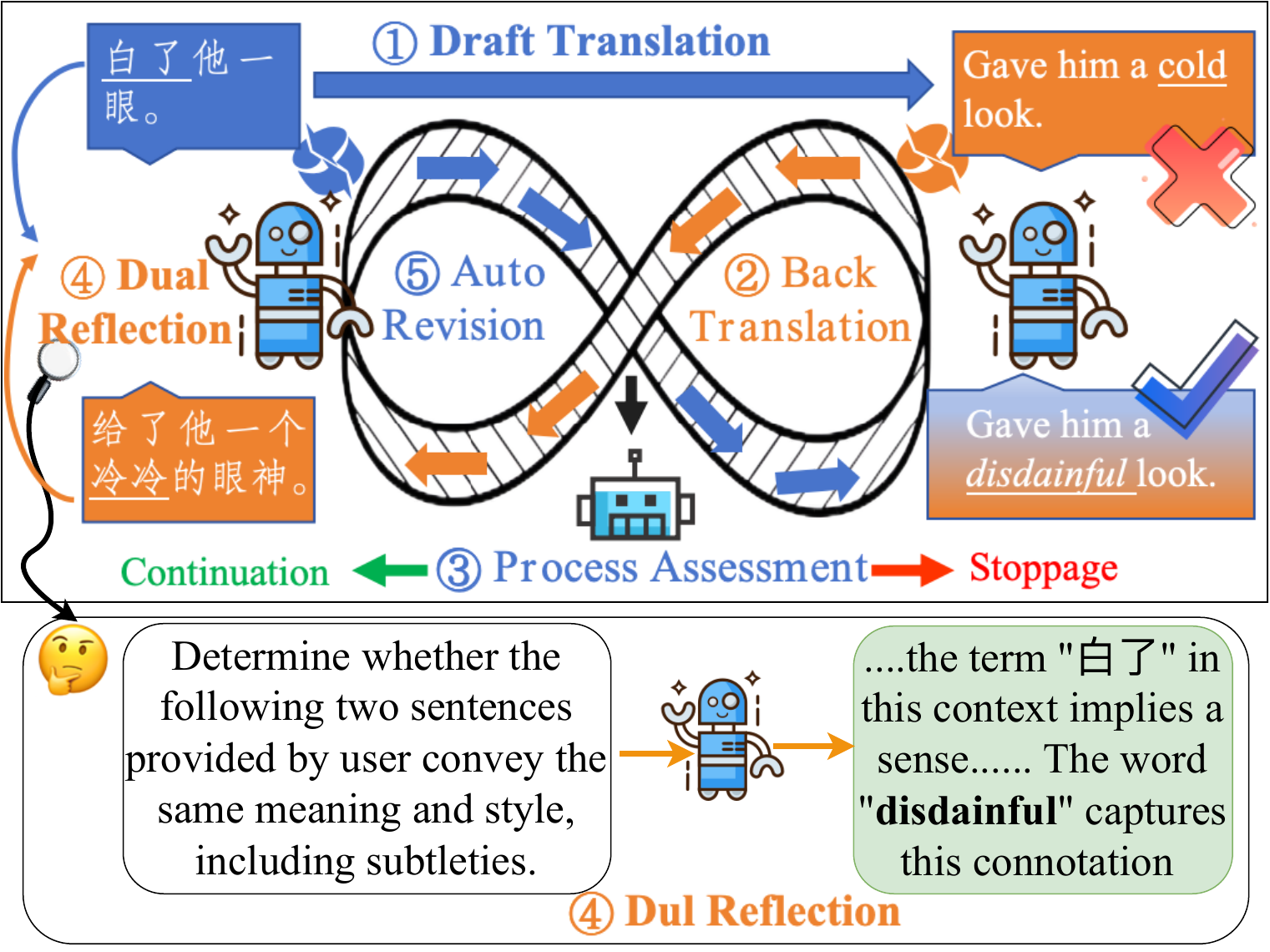} 
\caption{DUAL-REFLECT first obtains an initial translation result, then performs back-translation, and LLMs reflect on the differences between the back-translation results and the original source content to obtain feedback signals, ultimately optimizing the translation outcome.}
\label{fig.1}
\end{figure}

To address this, we introduce a framework that leverages the inherent duality property ~\cite{he2016dual,qin2020dual,sun2021tibetan,yi2017dualgan,xia2017dual} of translation tasks to provide effective feedback to LLMs, thereby enhancing their reflective capabilities and consequently improving translation performance. 
This method, named DUAL-REFLECT, stands for \textbf{DUAL} learning enhanced auto-\textbf{REFLEC}tive \textbf{T}ranslation and comprises five stages: Draft Translation, Back Translation, Process Assessment, Dual-Reflection, Auto Revision. 
In the draft translation stage, LLMs employ their inherent translation capabilities to generate a draft translation. 
Subsequently, in the Back Translation stage, LLMs translate the draft translation back to the source language. 
Then, during the process assessment stage, an LLM-based agent is introduced to assess whether dual reflection is needed. If not, it outputs the final result; otherwise, the process continues to cycle through all the steps.
Based on this, in the dual reflection stage, LLMs reflect on the differences between the back-translation results and the initial source input, revealing potential translation biases. LLMs further analyze the reasons for these discrepancies and propose suggestions for improvement. 
Finally, In the auto-revision stage, LLMs modify the initial translation by incorporating the analysis and improvement suggestions obtained through dual reflection.

We verify the effectiveness of the DUAL-REFLECT framework across four translation directions in the WMT22, covering high, medium, and lower resource languages, as well as a commonsense reasoning MT Benchmark.
Automatic evaluation results show that DUAL-REFLECT outperforms strong baseline methods, significantly enhancing translation performance.
Notably, on low-resource translation tasks, DUAL-REFLECT achieved an average result that surpassed ChatGPT by +1.6 COMET.
In addition, DUAL-REFLECT enhanced ChatGPT exceeded GPT-4 on the commonsense reasoning MT benchmark.
Further human evaluation demonstrates that DUAL-REFLECT shows a better ability to resolve translation ambiguities compared to other methods.

\section{Approach: DUAL-REFLECT}
Our DUAL-REFLECT framework consists of Five key stages, described in detail as follows:

\subsection{Stage-1: Draft Translation}
In the draft translation stage, LLMs utilize their inherent translation capabilities to generate a draft translation from the source language \(L^s\) to the target language \(L^t\). The instruction template for this translation task is as follows:
\begin{mdframed}[backgroundcolor=purple!10, linecolor=white, linewidth=2pt, roundcorner=10pt]\small
\textbf{Translation Instruction:} Translate the following text from \(L^s\) to \(L^t\):

\textbf{Input Text}: \begin{mdframed}[backgroundcolor=blue!10, linecolor=purple!10, linewidth=2pt, roundcorner=10pt]Source Sentence \(x\)\end{mdframed}\small

\textbf{Output Text}: \begin{mdframed}[backgroundcolor=yellow!10, linecolor=purple!10, linewidth=2pt, roundcorner=10pt]Target Sentence $y$\end{mdframed}\small
\end{mdframed}

\subsection{Stage-2: Back Translation}
In this stage, the same instruction as used in the draft translation stage is adopted. The goal is to back-translate the initial translation result from the target language \(L^t\) back to the source language \(L^s\), with the output being \(x'\).

\subsection{Stage-3: Process Assessment}
\label{section2.5}
We introduce an evaluation agent, denoted as \(PA\), to supervise and control the entire translation process. This Agent has two different modes:

\textbf{Judgment Mode}: \(PA\) determines whether it can accurately identify the differences between $x$ and $x^{\prime}$ within a given specific number of iterations. If $PA(x,x^{\prime}) = False$, the Dual Reflection stage is terminated; otherwise, the entire process continues.

\label{detail stage3}
\begin{mdframed}[backgroundcolor=purple!10, linecolor=white, linewidth=2pt, roundcorner=10pt]\small
\textbf{Stage-3: Judgment Mode:} If you are a $L^s$ linguist, Determine whether the following two sentences provided by user convey the same meaning and style, including subtleties. If so, give '$False$' response without any explanation, otherwise give '$True$' response and explain the reason.

\textbf{Input Text}: \begin{mdframed}[backgroundcolor=blue!10, linecolor=purple!10, linewidth=2pt, roundcorner=10pt]Source Sentence \(x\) and Back Translation Output $x^{\prime}$\end{mdframed}\small

\textbf{Output Text}: \begin{mdframed}[backgroundcolor=yellow!10, linecolor=purple!10, linewidth=2pt, roundcorner=10pt]'$True$' or '$False$'\end{mdframed}\small
\end{mdframed}


\textbf{Pattern Extraction}: In the judgment mode, once determined to be $True$ or after exceeding the predefined number of iterations, \(PA\) is responsible for extracting the final translation result from the entire output, denoted as \(PA(x,x^{\prime}) = final\_translation\).

\begin{mdframed}[backgroundcolor=purple!10, linecolor=white, linewidth=2pt, roundcorner=10pt]\small
\textbf{Stage-3: Pattern Extraction:} Therefore, $Pattern$ $Extraction:$ Please summarize the input information, you need to extract the final translation result from the paragraph.  Now, please output your answer in JSON format, as follows: \\
$\{'final\_translation': ''\}$. Please strictly follow the JSON format and do not output irrelevant content.

\textbf{Input Text}: \begin{mdframed}[backgroundcolor=blue!10, linecolor=purple!10, linewidth=2pt, roundcorner=10pt]Target Sentence $y$\end{mdframed}\small

\textbf{Output Text}: \begin{mdframed}[backgroundcolor=yellow!10, linecolor=purple!10, linewidth=2pt, roundcorner=10pt]\{'final\_translation': 'extraction result'\}\end{mdframed}\small
\end{mdframed}

\subsection{Stage-4: Dual Reflection}
The goal of the dual reflection stage is to reflect on the differences between the source sentences generated by back-translation and the initial source input. Then, it outputs analysis results and proposes suggestions to enhance translation performance.
\begin{mdframed}[backgroundcolor=purple!10, linecolor=white, linewidth=2pt, roundcorner=10pt]\small
\textbf{Dual Reflection Instruction}: Compare the the two sentences provided by the user. It aims to analyze the disparities between them in meaning, style, and subtleties, first provide analytical results, and then suggest how to revise them to make the two sentences consistent.
\\
\textbf{Input Text}: \begin{mdframed}[backgroundcolor=blue!10, linecolor=purple!10, linewidth=2pt, roundcorner=10pt]Source Sentence $x^{\prime}$ and $x$\end{mdframed}\small

\textbf{Output Text}: \begin{mdframed}[backgroundcolor=yellow!10, linecolor=purple!10, linewidth=2pt, roundcorner=10pt]Analysis Results ($AR$) and Translation Suggestions ($TS$)\end{mdframed}\small
\end{mdframed}

\subsection{Stage-5: Auto Revision}
In this stage, utilizing the output of the dual reflection and the original source sentences as input, the original source sentences are re-translated (from \(L^s\) to \(L^t\)).
\begin{mdframed}[backgroundcolor=purple!10, linecolor=white, linewidth=2pt, roundcorner=10pt]\small
\textbf{Auto Revision Instruction}: Translate the following text from \(L^s\) to \(L^t\):

\textbf{Input Text}: \begin{mdframed}[backgroundcolor=blue!10, linecolor=purple!10, linewidth=2pt, roundcorner=10pt]Analysis Results ($AR$), Translation Suggestions ($TS$) and $x$\end{mdframed}

\textbf{Output Text}: \begin{mdframed}[backgroundcolor=yellow!10, linecolor=purple!10, linewidth=2pt, roundcorner=10pt]Target Sentence $y$\end{mdframed}
\end{mdframed}

\begin{table*}[]\small
\begin{tabular}{lcccccccc}
\hline
\multirow{2}{*}{Methods} & \multicolumn{2}{c}{En$\rightarrow$De} & \multicolumn{2}{c}{En$\rightarrow$Ja} & \multicolumn{2}{c}{Cs$\rightarrow$Uk} & \multicolumn{2}{c}{En$\rightarrow$Hr} \\ \cline{2-9} 
                         & COMET             & BLEURT            & COMET             & BLEURT            & COMET             & BLEURT            & COMET             & BLEURT            \\ \hline
ChatGPT                  & 85.8              & 75.6              & 87.9              & 66.3              & 88.0              & 75.0              & 85.9              & 75.0              \\
\quad+5-shot                  & 86.5              & 76.3              & 88.2              & 67.1              & 88.3              & -                 & 86.4              & -                 \\
\quad+Rerank                  & 86.0              & 75.9              & 88.0              & 66.6              & 88.3              & 75.3              & 86.3              & 75.4              \\
\quad+Refine                  & 85.9              & 76.0              & 88.1              & 66.4              & 89.0              & 74.5              & 86.1              & 75.6              \\
\quad+Refine\_cos                  & 86.2              & 76.3              & 88.4              & 66.8              & 89.5              & 75.0              & 86.4              & 75.9              \\
\quad+MAPS                    & 86.4              & 76.3              & 88.5              & 67.4              & 88.8              & 76.1              & 86.5              & 76.0              \\
\quad+Self-Reflect            & 86.3              & 76.1              & 88.3              & 66.9              & 88.4              & 76.0              & 86.3              & 75.8              \\
\quad+DUAL-REFLECT            & \textbf{86.5}     & \textbf{76.4}     & \textbf{88.7}     & \textbf{67.9}     & \textbf{90.2}     & \textbf{77.3}     & \textbf{86.9}     & \textbf{76.4}     \\ \hline
Alpaca-7B                & 75.5              & 62.2
                  & 56.6              &  31.4
                 & 74.1              &     52.4
              & 65.9              &    53.2
               \\
\quad+5shot                   & 76.3              &  62.8
                 & 57.9              &   31.9                & 75.9              &    53.1

                & 67.9              &  53.6
                 \\
\quad+MAPS                    & 76.7              &  63.5
                 & 58.2              & 33.9
                  & 76.3              &   53.7
                & 68.1              &  54.2
                 \\
\quad+DUAL-REFLECT         & \textbf{78.1}     & \textbf{64.1}         & \textbf{61.0}     & \textbf{34.7}         & \textbf{77.5}     & \textbf{54.3}         & \textbf{69.5}     &     \textbf{55.4}
              \\ \hline
Vicuna-7B                & 79.8              &  67.4
                 & 82.3              &     58.7
              & 74.9              & 57.8
                  & 69.3              &   57.7
                \\
\quad+5shot                   & 80.3              &      67.8
             & 83.3              &  59.3
                 & 76.3              &  58.3
                 & 70.2              &   58.1
                \\
\quad+MAPS                    & 81.1              & 68.4
                  & 84.4              &  60.3
                 & 77.2              &    59.6
               & 71.1              &     58.8
              \\
\quad+DUAL-REFLECT         & \textbf{82.0}     & \textbf{69.1}         & \textbf{85.1}     & \textbf{61.1}         & \textbf{78.3}     & \textbf{60.7}         & \textbf{72.9}     &  \textbf{60.4}
                 \\ \hline
\end{tabular}
\caption{The main results from the WMT22 benchmark are presented. ChatGPT, Alpaca-7B, and Vicuna-7B mean to perform translation directly through Zero-Shot. The bold indicates the highest values that are statistically significant, with p-values less than 0.05 in the paired t-test against all compared methods.}
\label{wmt22test}
\end{table*}

\section{Experiments}
\subsection{Experimental Setup}

\textbf{Test Data.} To mitigate concerns of data leakage as highlighted by ~\citealp{DBLP:journals/corr/abs-2303-12712}, ~\citealp{garcia2023unreasonable}, and ~\citealp{zhu2023multilingual}, we leveraged the WMT22\footnote{https://www.statmt.org/wmt22/index.html} ~\cite{kocmi2022findings} and WMT23\footnote{https://www2.statmt.org/wmt23/} ~\cite{kocmi2023findings} test set in our evaluation framework. Additionally, to further evaluate DUAL-REFLECT's performance in complex translation tasks, we employed the Commonsense Reasoning MT dataset \cite{he2020box}, consisting of Chinese$\rightarrow$English translation examples. See Appendix \ref{sec:app_datasets} for specific details.

\noindent\textbf{Comparing Systems.} In our evaluation, the DUAL-REFLECT framework is compared with a range of models, including ChatGPT \cite{ouyang2022training}, GPT-4\footnote{The ChatGPT and GPT-4 models used in this work are accessed through the gpt-3.5-turbo and gpt-4 APIs, respectively.} \cite{achiam2023gpt}, Alpaca-7B\footnote{https://huggingface.co/tatsu-lab/alpaca-7b-wdiff/tree/main}, Vicuna-7B\footnote{https://huggingface.co/lmsys/vicuna-7b-v1.5}, ReRank~\cite{He2023ExploringHT}, Self-Reflect \cite{shinn2023reflexion}, MAD \cite{liang2023encouraging}, and MAPS \cite{He2023ExploringHT}. See Appendix \ref{sec:Comparative_Methods} for specific details.

\noindent\textbf{Evaluation Metrics.} In evaluating our translation methodology, we initially employ COMET\footnote{https://huggingface.co/Unbabel/wmt22-comet-da} \cite{rei2022comet} and BLEURT\footnote{https://github.com/lucadiliello/bleurt-pytorch} \cite{sellam2020bleurt} as automatic metrics, aligning with the established standards in LLM-based translation literature \cite{He2023ExploringHT,huang2024aligning}. To further evaluate our translation method, we employ human evaluations to verify translation performance and the ability to resolve translation ambiguities. Details on human evaluations are in Appendix \ref{sec:human_evaluation}.


\subsection{Main Results}
The main results of WMT22 and the Commonsense MT are presented in Tables \ref{wmt22test} and \ref{commenmt}. The results of WMT23 are presented in Appendix \ref{sec:wmt23}. Based on these outcomes, we derive the subsequent insights:

\textbf{The effectiveness of DUAL-REFLECT has been validated across a wide range of settings.} As shown in Table \ref{wmt22test}, across 4 language pairs, 3 LLMs, and 2 metrics, DUAL-REFLECT achieves the best performance compared to other methods. Specifically, DUAL-REFLECT demonstrates an average improvement of +1.18 COMET over the baseline ChatGPT and +0.75 COMET over the Self-Reflect methods. In the low-resource Cs$\rightarrow$Uk translation task, DUAL-REFLECT surpasses ChatGPT and MAPS by +2.2 and +1.4 COMET, respectively. 
Additionally, Table \ref{wmt22_add_low} shows the remaining five low-resource tasks from WMT22, with an average increase of +0.7 COMET.
These improvements indicate that DUAL-REFLECT has broad applicability across different levels of resource availability and language similarity, especially exhibiting more pronounced improvements in language pairs with lower resources.



\textbf{The effectiveness of DUAL-REFLECT in commonsense reasoning translation tasks.} The results, presented in Table \ref{commenmt}, show that in commonsense reasoning translation tasks, DUAL-REFLECT significantly outperforms other methods, achieving the best translation performance. Compared to the Self-Reflect method, it showed an improvement of +1.3 COMET, indicating more effective error correction capabilities. Moreover, DUAL-REFLECT also surpassed the MAD method, which relies on feedback from multi-agent debate, demonstrating the high quality of its feedback. Notably, in translation tasks involving logical reasoning, DUAL-REFLECT's performance even exceeded that of GPT-4, suggesting reasoning abilities.

\begin{table}[!h]\small
\centering
\begin{tabular}{lcc}
\hline
\multirow{2}{*}{\textbf{Methods}}       & \multicolumn{2}{c}{\textbf{AutoMetrics}}             \\ \cline{2-3} 
                               & \textbf{COMET}                 & \textbf{BLEURT}                 \\ \hline
\textbf{GPT-4}                          & 82.0                 & 71.0                    \\ \hline
\textbf{ChatGPT} & \multicolumn{1}{l}{} & \multicolumn{1}{l}{} \\
 \quad\textbf{+Zero-Shot}                      & 79.7                & 68.2                    \\
 \quad\textbf{+Rerank}                         & 80.9                 & 68.9                    \\
  \quad\textbf{+Refine}                         & 80.4                 & 68.5                    \\
 \quad\textbf{+Refine\_cos}                     & 80.8                 & 68.8                    \\
 \quad\textbf{+MAPS}                           & 81.9                 & -                    \\
 \quad\textbf{+Self-Reflect}                   & 80.9                 & 68.7                    \\
 \quad\textbf{+MAD}                            & 82.0                 & 69.4                    \\
 \quad\textbf{+DUAL-REFLECT}                    & \textbf{82.2}                 & \textbf{71.8}                    \\ \hline
\end{tabular}
\caption{The main results from the Commonsense MT benchmark are presented.  The bold indicates the highest value. The bold indicates the highest values, statistically significant with p-values less than 0.05 in the paired t-test against compared methods.}
\label{commenmt}
\end{table}

\vspace{-0.33 cm}

\section{Analysis}
We thoroughly analyze our approach, with results primarily reported on CommonsenseMT Zh$\rightarrow$En unless stated otherwise.

\subsection{The Effectiveness of Dual Learning}
\begin{figure}[!th]\centering
\centering
\includegraphics[scale=0.28]{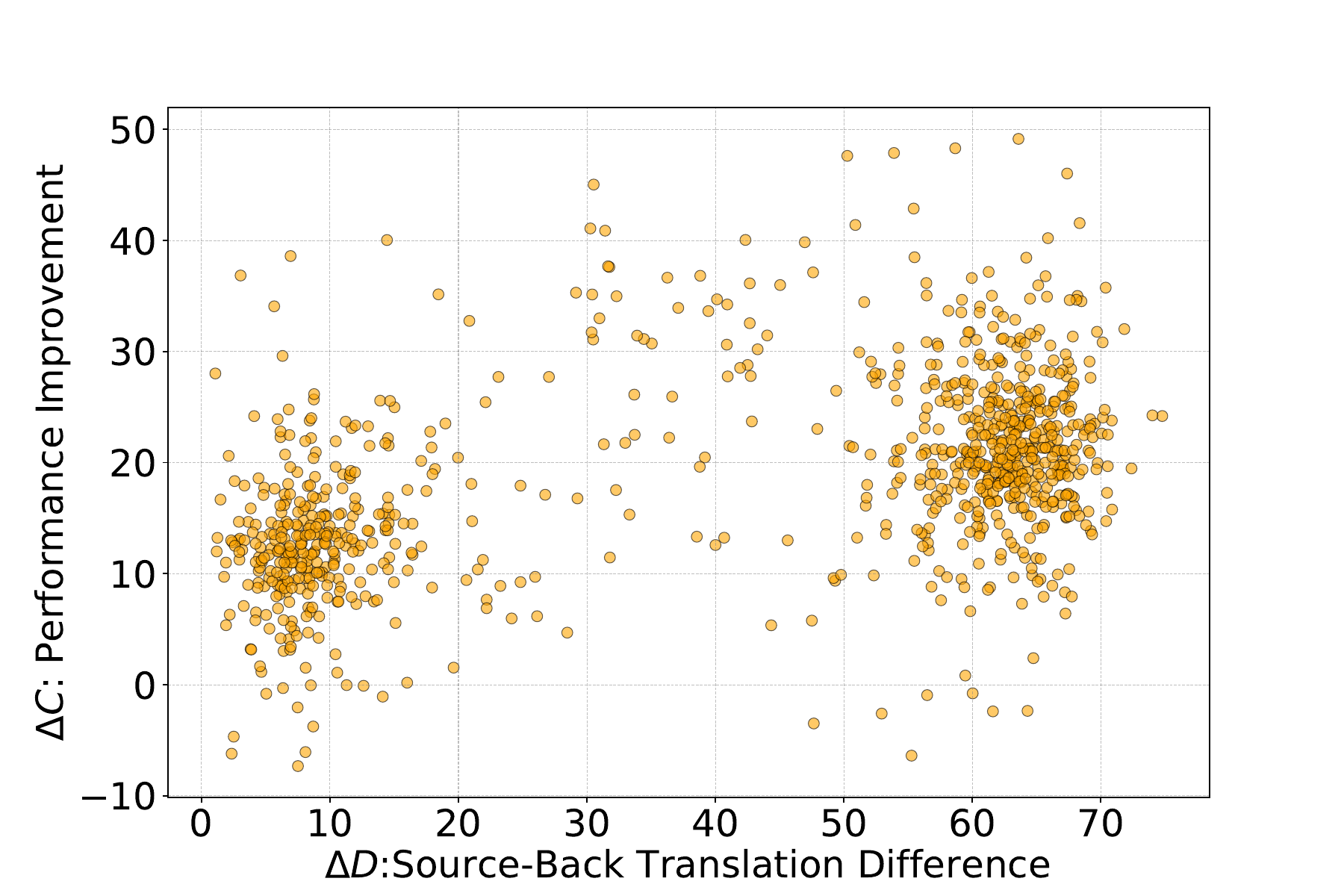} 
\caption{Effectiveness experiment of Dual Learning, each point represents a translation data from the test set.}
\label{dul}
\end{figure}

In this study, we explore the potential positive impact of a dual learning feedback mechanism on translation performance, as shown in Figure \ref{dul}. The horizontal axis denotes $\Delta D=100-COMET(x,x^{\prime})$, the disparity between the original sentence $x$ and its back-translated version $x^{\prime}$. The vertical axis quantifies improvement in translation performance, as a COMET metric difference ($\Delta C$), between DUAL-REFLECT and ChatGPT. Findings show a correlation coefficient of 0.46, indicating that feedback from dual learning improves the model's reflective capabilities, thus enhancing translation accuracy. Additionally, the experimental data shows significant differences between the output $x^{\prime}$ and the original source sentence $x$ in the initial back-translation ($\Delta D > 50$), further confirming the universality of differences obtained from the dual learning in translation tasks.
\vspace{-0.5 cm}

\subsection{Human Evaluation} 

In terms of human evaluation, this study follows the method of \citealp{liang2023encouraging} to assess translation outcomes from two main dimensions: accuracy in ambiguity resolution and direct assessment of translation quality (details in Appendix \ref{sec:human_evaluation}).

The experimental results are presented in Table \ref{human}. Regarding the accuracy of ambiguity resolution, DUAL-REFLECT performs the best, indicating that dual feedback contributes to better disambiguation in translation tasks. In terms of human evaluation, DUAL-REFLECT receives the highest ratings, further demonstrating that the method achieves superior translation quality.
\begin{table}[ht]\small
\centering
\begin{tabular}{lcc}
\hline
\multirow{2}{*}{\textbf{Methods}}       & \multicolumn{2}{c}{\textbf{Human Evaluation}}             \\ \cline{2-3} 
                               & \textbf{Score}                 & \textbf{ACC}                 \\ \hline
\textbf{GPT-4}                          & 3.9                 & 69.8                    \\ \hline
\textbf{ChatGPT} & \multicolumn{1}{l}{} & \multicolumn{1}{l}{} \\
 \quad\textbf{+Zero-Shot}                      & 3.1                & 63.8                    \\
 \quad\textbf{+Rerank}                         & 3.3                 & 66.8                                       \\
 \quad\textbf{+Self-Reflect}                   & 3.4                & 64.9                    \\
 \quad\textbf{+MAD}                            & 3.7                 & 76.2                    \\
 \quad\textbf{+DUAL-REFLECT}                    & \textbf{4.2}                 & \textbf{77.4}                    \\ \hline
\end{tabular}
\caption{The human-annotated results of the Commonsense MT benchmark.}
\label{human}
\end{table}
\vspace{-0.3cm}

\subsection{Examine how iteration rounds affect results} 
In this experimental design, we require reviewer $PA$ to determine the final answer (\(PA(x,x^{\prime}) = final\_translation\)) in each iteration, rather than allowing adaptive termination of iterations as described in Section \ref{section2.5}. Figure \ref{iter} in the Appendix presents the outcomes, revealing DUAL-REFLECT's superior performance over the benchmark method as iterations progress, notably achieving the highest COMET score in three iterations. This emphasizes DUAL-REFLECT's ability to provide improved translations through repeated iterations, demonstrating the effectiveness and robustness of its dual learning feedback mechanism.

\begin{figure}[!ht]\centering
\centering
\includegraphics[scale=0.033]{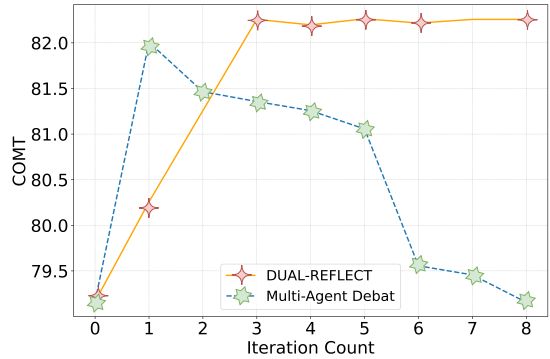} 
\caption{Translation performance varies based on the iteration of either DUAL-REFLECT or MAD.}
\label{iter}
\end{figure}


\section{Case Study}
\begin{figure}[!ht]\centering
\includegraphics[scale=0.4]{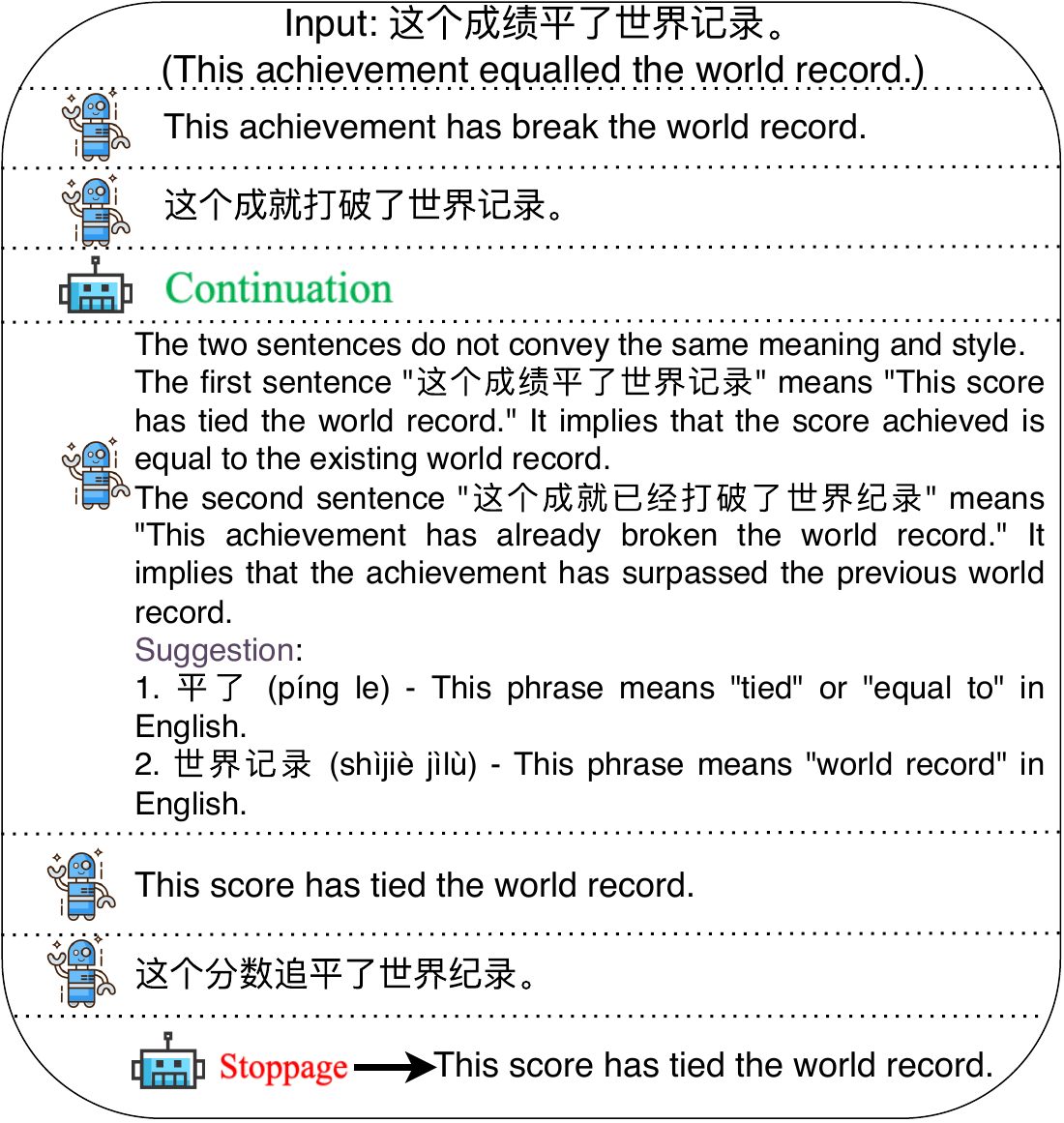} 
\caption{The DUAL-REFLECT methodology for translating \underline{positive} examples within Chinese sentences.}
\label{case2}
\end{figure}

This section presents a case study on the DUAL-REFLECT method, assessing its effectiveness and constraints via examples (detailed in Appendix \ref{sec:case_study}). Positive instances (Figures \ref{case2}, \ref{case1} and \ref{case3}) illustrate substantial enhancements in translation accuracy and semantic coherence due to its reflective and iterative processes. Conversely, negative examples (Figures \ref{case4} and \ref{case5}) highlight the dependency of DUAL-REFLECT's success on Back Translation quality, suggesting limitations in its capacity for improvement. This underscores the method's reliance on the integrity of each cycle component for optimal performance.

\vspace{-0.3cm}


\section{Conclusion}
We introduced DUAL-REFLECT, an LLM-based machine translation method, that leverages dual learning to improve reflection and performance, excelling in resource-limited and common sense reasoning scenarios, with human evaluations confirming its effectiveness.

\section{Limitations}
The DUAL-REFLECT framework enhances the reflective capabilities of LLMs in translation tasks by leveraging the duality nature of translation but has several limitations. Firstly, models with stronger reflective capabilities will obtain better feedback, thereby enhancing more performance. Additionally, since our method requires multiple steps, it necessitates a significant amount of computational resources.

\section{Ethics Statement}
One of the core design principles of the DUAL-REFLECT framework is a strict respect for intellectual property rights. This applies to both the methods and algorithms developed within the framework as well as those cited from the literature, all adhering strictly to copyright laws. Additionally, the framework upholds this principle in the handling of translation content, ensuring its use does not infringe upon the rights of original creators.

The framework also places a strong emphasis on responsibility during the automated translation process. By integrating stages of reflection and revision, DUAL-REFLECT enhances the transparency and interpretability of the translation methodology, thereby effectively identifying and correcting potential errors in the translation process.

\section{Acknowledgements}

We want to thank all the anonymous reviewers for their valuable comments. This work was supported by the National Natural Science Foundation of China (62276077, 62376075, U1908216, 62376076 and 62106115), Guangdong Basic and Applied Basic Research Foundation (2024A1515011205), Shenzhen College Stability Support Plan (GXWD20220811170358002, GXWD20220817123150002), Key R\&D Program of Yunnan (202203AA080004), and Major Key Project of PCL under Grant No. PCL2022D01.

\bibliography{anthology,custom}

\begin{thebibliography}{37}
\expandafter\ifx\csname natexlab\endcsname\relax\def\natexlab#1{#1}\fi

\bibitem[{Achiam et~al.(2023)Achiam, Adler, Agarwal, Ahmad, Akkaya, Aleman, Almeida, Altenschmidt, Altman, Anadkat et~al.}]{achiam2023gpt}
Josh Achiam, Steven Adler, Sandhini Agarwal, Lama Ahmad, Ilge Akkaya, Florencia~Leoni Aleman, Diogo Almeida, Janko Altenschmidt, Sam Altman, Shyamal Anadkat, et~al. 2023.
\newblock Gpt-4 technical report.
\newblock \emph{arXiv preprint arXiv:2303.08774}.

\bibitem[{Amabile(1983)}]{amabile1983theoretical}
Teresa~M Amabile. 1983.
\newblock A theoretical framework.
\newblock \emph{The Social Psychology of Creativity}, pages 65--96.

\bibitem[{Bubeck et~al.(2023{\natexlab{a}})Bubeck, Chandrasekaran, Eldan, Gehrke, Horvitz, Kamar, Lee, Lee, Li, Lundberg et~al.}]{bubeck2023sparks}
S{\'e}bastien Bubeck, Varun Chandrasekaran, Ronen Eldan, Johannes Gehrke, Eric Horvitz, Ece Kamar, Peter Lee, Yin~Tat Lee, Yuanzhi Li, Scott Lundberg, et~al. 2023{\natexlab{a}}.
\newblock Sparks of artificial general intelligence: Early experiments with gpt-4.
\newblock \emph{arXiv preprint arXiv:2303.12712}.

\bibitem[{Bubeck et~al.(2023{\natexlab{b}})Bubeck, Chandrasekaran, Eldan, Gehrke, Horvitz, Kamar, Lee, Lee, Li, Lundberg, Nori, Palangi, Ribeiro, and Zhang}]{DBLP:journals/corr/abs-2303-12712}
S{\'{e}}bastien Bubeck, Varun Chandrasekaran, Ronen Eldan, Johannes Gehrke, Eric Horvitz, Ece Kamar, Peter Lee, Yin~Tat Lee, Yuanzhi Li, Scott~M. Lundberg, Harsha Nori, Hamid Palangi, Marco~T{\'{u}}lio Ribeiro, and Yi~Zhang. 2023{\natexlab{b}}.
\newblock Sparks of artificial general intelligence: Early experiments with {GPT-4}.
\newblock \emph{CoRR}.

\bibitem[{Chen et~al.(2023{\natexlab{a}})Chen, Sun, Zhao, Galindo~Esparza, Chen, Xiang, Zhao, and Zhang}]{chen-etal-2023-improving-low}
Andong Chen, Yuan Sun, Xiaobing Zhao, Rosella Galindo~Esparza, Kehai Chen, Yang Xiang, Tiejun Zhao, and Min Zhang. 2023{\natexlab{a}}.
\newblock Improving low-resource question answering by augmenting question information.
\newblock In \emph{Findings of the Association for Computational Linguistics: EMNLP 2023}, pages 10413--10420, Singapore. Association for Computational Linguistics.

\bibitem[{Chen et~al.(2023{\natexlab{b}})Chen, Yao, Zhao, Zhang, Sun, Liu, and Shen}]{chen2023equals}
Andong Chen, Feng Yao, Xinyan Zhao, Yating Zhang, Changlong Sun, Yun Liu, and Weixing Shen. 2023{\natexlab{b}}.
\newblock Equals: A real-world dataset for legal question answering via reading chinese laws.
\newblock In \emph{Proceedings of the Nineteenth International Conference on Artificial Intelligence and Law}, pages 71--80.

\bibitem[{Chen et~al.(2023{\natexlab{c}})Chen, Guo, Haddow, and Heafield}]{chen2023iterative}
Pinzhen Chen, Zhicheng Guo, Barry Haddow, and Kenneth Heafield. 2023{\natexlab{c}}.
\newblock Iterative translation refinement with large language models.
\newblock \emph{arXiv preprint arXiv:2306.03856}.

\bibitem[{Flower and Hayes(1981)}]{flower1981cognitive}
Linda Flower and John~R Hayes. 1981.
\newblock A cognitive process theory of writing.
\newblock \emph{College composition and communication}, 32(4):365--387.

\bibitem[{Garcia et~al.(2023)Garcia, Bansal, Cherry, Foster, Krikun, Johnson, and Firat}]{garcia2023unreasonable}
Xavier Garcia, Yamini Bansal, Colin Cherry, George Foster, Maxim Krikun, Melvin Johnson, and Orhan Firat. 2023.
\newblock The unreasonable effectiveness of few-shot learning for machine translation.
\newblock In \emph{International Conference on Machine Learning}, pages 10867--10878. PMLR.

\bibitem[{He et~al.(2016)He, Xia, Qin, Wang, Yu, Liu, and Ma}]{he2016dual}
Di~He, Yingce Xia, Tao Qin, Liwei Wang, Nenghai Yu, Tie-Yan Liu, and Wei-Ying Ma. 2016.
\newblock Dual learning for machine translation.
\newblock \emph{Advances in neural information processing systems}, 29.

\bibitem[{He et~al.(2020)He, Wang, Xiong, and Liu}]{he2020box}
Jie He, Tao Wang, Deyi Xiong, and Qun Liu. 2020.
\newblock The box is in the pen: Evaluating commonsense reasoning in neural machine translation.
\newblock In \emph{Findings of the Association for Computational Linguistics: EMNLP 2020}, pages 3662--3672.

\bibitem[{He et~al.(2023)He, Liang, Jiao, Zhang, Yang, Wang, Tu, Shi, and Wang}]{He2023ExploringHT}
Zhiwei He, Tian Liang, Wenxiang Jiao, Zhuosheng Zhang, Yujiu Yang, Rui Wang, Zhaopeng Tu, Shuming Shi, and Xing Wang. 2023.
\newblock Exploring human-like translation strategy with large language models.
\newblock \emph{ArXiv}, abs/2305.04118.

\bibitem[{Huang et~al.(2024)Huang, Feng, Li, Fu, Huo, Liu, and Qin}]{huang2024aligning}
Yichong Huang, Xiaocheng Feng, Baohang Li, Chengpeng Fu, Wenshuai Huo, Ting Liu, and Bing Qin. 2024.
\newblock Aligning translation-specific understanding to general understanding in large language models.
\newblock \emph{arXiv preprint arXiv:2401.05072}.

\bibitem[{Kocmi et~al.(2023)Kocmi, Avramidis, Bawden, Bojar, Dvorkovich, Federmann, Fishel, Freitag, Gowda, Grundkiewicz et~al.}]{kocmi2023findings}
Tom Kocmi, Eleftherios Avramidis, Rachel Bawden, Ond{\v{r}}ej Bojar, Anton Dvorkovich, Christian Federmann, Mark Fishel, Markus Freitag, Thamme Gowda, Roman Grundkiewicz, et~al. 2023.
\newblock Findings of the 2023 conference on machine translation (wmt23): Llms are here but not quite there yet.
\newblock In \emph{Proceedings of the Eighth Conference on Machine Translation}, pages 1--42.

\bibitem[{Kocmi et~al.(2022)Kocmi, Bawden, Bojar, Dvorkovich, Federmann, Fishel, Gowda, Graham, Grundkiewicz, Haddow et~al.}]{kocmi2022findings}
Tom Kocmi, Rachel Bawden, Ond{\v{r}}ej Bojar, Anton Dvorkovich, Christian Federmann, Mark Fishel, Thamme Gowda, Yvette Graham, Roman Grundkiewicz, Barry Haddow, et~al. 2022.
\newblock Findings of the 2022 conference on machine translation (wmt22).
\newblock In \emph{Proceedings of the Seventh Conference on Machine Translation (WMT)}, pages 1--45.

\bibitem[{Le et~al.(2022)Le, Wang, Gotmare, Savarese, and Hoi}]{le2022coderl}
Hung Le, Yue Wang, Akhilesh~Deepak Gotmare, Silvio Savarese, and Steven Chu~Hong Hoi. 2022.
\newblock Coderl: Mastering code generation through pretrained models and deep reinforcement learning.
\newblock \emph{Advances in Neural Information Processing Systems}, 35:21314--21328.

\bibitem[{Liang et~al.(2023)Liang, He, Jiao, Wang, Wang, Wang, Yang, Tu, and Shi}]{liang2023encouraging}
Tian Liang, Zhiwei He, Wenxiang Jiao, Xing Wang, Yan Wang, Rui Wang, Yujiu Yang, Zhaopeng Tu, and Shuming Shi. 2023.
\newblock Encouraging divergent thinking in large language models through multi-agent debate.
\newblock \emph{arXiv preprint arXiv:2305.19118}.

\bibitem[{Lou et~al.(2023)Lou, Yin, Xie, and Xiang}]{lou-etal-2023-cceval}
Lianzhang Lou, Xi~Yin, Yutao Xie, and Yang Xiang. 2023.
\newblock {CCE}val: A representative evaluation benchmark for the {C}hinese-centric multilingual machine translation.
\newblock In \emph{Findings of the Association for Computational Linguistics: EMNLP 2023}, pages 10176--10184, Singapore. Association for Computational Linguistics.

\bibitem[{Moslem et~al.(2023)Moslem, Haque, and Way}]{moslem2023adaptive}
Yasmin Moslem, Rejwanul Haque, and Andy Way. 2023.
\newblock Adaptive machine translation with large language models.
\newblock \emph{arXiv preprint arXiv:2301.13294}.

\bibitem[{Ouyang et~al.(2022)Ouyang, Wu, Jiang, Almeida, Wainwright, Mishkin, Zhang, Agarwal, Slama, Ray et~al.}]{ouyang2022training}
Long Ouyang, Jeffrey Wu, Xu~Jiang, Diogo Almeida, Carroll Wainwright, Pamela Mishkin, Chong Zhang, Sandhini Agarwal, Katarina Slama, Alex Ray, et~al. 2022.
\newblock Training language models to follow instructions with human feedback.
\newblock \emph{Advances in Neural Information Processing Systems}, 35:27730--27744.

\bibitem[{Park et~al.(2023)Park, O'Brien, Cai, Morris, Liang, and Bernstein}]{DBLP:conf/uist/ParkOCMLB23}
Joon~Sung Park, Joseph~C. O'Brien, Carrie~Jun Cai, Meredith~Ringel Morris, Percy Liang, and Michael~S. Bernstein. 2023.
\newblock Generative agents: Interactive simulacra of human behavior.
\newblock In \emph{Proceedings of the 36th Annual {ACM} Symposium on User Interface Software and Technology, {UIST} 2023, San Francisco, CA, USA, 29 October 2023- 1 November 2023}, pages 2:1--2:22. {ACM}.

\bibitem[{Qin(2020)}]{qin2020dual}
Tao Qin. 2020.
\newblock \emph{Dual learning}.
\newblock Springer.

\bibitem[{Rei et~al.(2022{\natexlab{a}})Rei, De~Souza, Alves, Zerva, Farinha, Glushkova, Lavie, Coheur, and Martins}]{rei2022comet}
Ricardo Rei, Jos{\'e}~GC De~Souza, Duarte Alves, Chrysoula Zerva, Ana~C Farinha, Taisiya Glushkova, Alon Lavie, Luisa Coheur, and Andr{\'e}~FT Martins. 2022{\natexlab{a}}.
\newblock Comet-22: Unbabel-ist 2022 submission for the metrics shared task.
\newblock In \emph{Proceedings of the Seventh Conference on Machine Translation (WMT)}, pages 578--585.

\bibitem[{Rei et~al.(2022{\natexlab{b}})Rei, Treviso, Guerreiro, Zerva, Farinha, Maroti, De~Souza, Glushkova, Alves, Lavie et~al.}]{rei2022cometkiwi}
Ricardo Rei, Marcos Treviso, Nuno~M Guerreiro, Chrysoula Zerva, Ana~C Farinha, Christine Maroti, Jos{\'e}~GC De~Souza, Taisiya Glushkova, Duarte~M Alves, Alon Lavie, et~al. 2022{\natexlab{b}}.
\newblock Cometkiwi: Ist-unbabel 2022 submission for the quality estimation shared task.
\newblock \emph{arXiv preprint arXiv:2209.06243}.

\bibitem[{Scheurer et~al.(2022)Scheurer, Campos, Chan, Chen, Cho, and Perez}]{scheurer2022training}
J{\'e}r{\'e}my Scheurer, Jon~Ander Campos, Jun~Shern Chan, Angelica Chen, Kyunghyun Cho, and Ethan Perez. 2022.
\newblock Training language models with natural language feedback.
\newblock \emph{arXiv preprint arXiv:2204.14146}, 8.

\bibitem[{Sellam et~al.(2020)Sellam, Das, and Parikh}]{sellam2020bleurt}
Thibault Sellam, Dipanjan Das, and Ankur~P Parikh. 2020.
\newblock Bleurt: Learning robust metrics for text generation.
\newblock \emph{arXiv preprint arXiv:2004.04696}.

\bibitem[{Shinn et~al.(2023)Shinn, Cassano, Berman, Gopinath, Narasimhan, and Yao}]{shinn2023reflexion}
Noah Shinn, Federico Cassano, Edward Berman, Ashwin Gopinath, Karthik Narasimhan, and Shunyu Yao. 2023.
\newblock Reflexion: Language agents with verbal reinforcement learning.

\bibitem[{Simon(1962)}]{simon1962architecture}
Herbert~A Simon. 1962.
\newblock The architecture of complexity.
\newblock \emph{Proceedings of the American philosophical society}, 106(6):467--482.

\bibitem[{Sun et~al.(2021{\natexlab{a}})Sun, Chen, Chen, Xia, and Zhao}]{sun2021joint}
Yuan Sun, Andong Chen, Chaofan Chen, Tianci Xia, and Xiaobing Zhao. 2021{\natexlab{a}}.
\newblock A joint model for representation learning of tibetan knowledge graph based on encyclopedia.
\newblock \emph{Transactions on Asian and Low-Resource Language Information Processing}, 20(2):1--17.

\bibitem[{Sun et~al.(2021{\natexlab{b}})Sun, Chen, Chen, and Zhao}]{sun2021tibetan}
Yuan Sun, Chaofan Chen, Andong Chen, and Xiaobing Zhao. 2021{\natexlab{b}}.
\newblock Tibetan question generation based on sequence to sequence model.
\newblock \emph{Computers, Materials \& Continua}, 68(3).

\bibitem[{Tyen et~al.(2023)Tyen, Mansoor, Chen, Mak, and C{\u{a}}rbune}]{tyen2023llms}
Gladys Tyen, Hassan Mansoor, Peter Chen, Tony Mak, and Victor C{\u{a}}rbune. 2023.
\newblock Llms cannot find reasoning errors, but can correct them!
\newblock \emph{arXiv preprint arXiv:2311.08516}.

\bibitem[{Welleck et~al.(2022)Welleck, Lu, West, Brahman, Shen, Khashabi, and Choi}]{welleck2022generating}
Sean Welleck, Ximing Lu, Peter West, Faeze Brahman, Tianxiao Shen, Daniel Khashabi, and Yejin Choi. 2022.
\newblock Generating sequences by learning to self-correct.
\newblock \emph{arXiv preprint arXiv:2211.00053}.

\bibitem[{Xia et~al.(2017)Xia, Qin, Chen, Bian, Yu, and Liu}]{xia2017dual}
Yingce Xia, Tao Qin, Wei Chen, Jiang Bian, Nenghai Yu, and Tie-Yan Liu. 2017.
\newblock Dual supervised learning.
\newblock In \emph{International conference on machine learning}, pages 3789--3798. PMLR.

\bibitem[{Xu and Poo(2023)}]{xu2023large}
Bo~Xu and Mu-ming Poo. 2023.
\newblock Large language models and brain-inspired general intelligence.
\newblock \emph{National Science Review}, 10(10):nwad267.

\bibitem[{Yi et~al.(2017)Yi, Zhang, Tan, and Gong}]{yi2017dualgan}
Zili Yi, Hao Zhang, Ping Tan, and Minglun Gong. 2017.
\newblock Dualgan: Unsupervised dual learning for image-to-image translation.
\newblock In \emph{Proceedings of the IEEE international conference on computer vision}, pages 2849--2857.

\bibitem[{Zhao et~al.(2023)Zhao, Zhou, Li, Tang, Wang, Hou, Min, Zhang, Zhang, Dong et~al.}]{zhao2023survey}
Wayne~Xin Zhao, Kun Zhou, Junyi Li, Tianyi Tang, Xiaolei Wang, Yupeng Hou, Yingqian Min, Beichen Zhang, Junjie Zhang, Zican Dong, et~al. 2023.
\newblock A survey of large language models.
\newblock \emph{arXiv preprint arXiv:2303.18223}.

\bibitem[{Zhu et~al.(2023)Zhu, Liu, Dong, Xu, Kong, Chen, Li, and Huang}]{zhu2023multilingual}
Wenhao Zhu, Hongyi Liu, Qingxiu Dong, Jingjing Xu, Lingpeng Kong, Jiajun Chen, Lei Li, and Shujian Huang. 2023.
\newblock Multilingual machine translation with large language models: Empirical results and analysis.
\newblock \emph{arXiv preprint arXiv:2304.04675}.

\end{thebibliography}

\appendix


\section{Experiment Setup}
\label{sec:Experiment_Setup}

\subsection{Test Data}
\label{sec:app_datasets}
For the WMT22 test set \cite{kocmi2022findings}, the experimental analysis covers 9 language pairs. We used the full test dataset. Among these languages, En$\rightarrow$De and En$\rightarrow$Ja are classified as high-resource and medium-resource languages, respectively. In contrast, Cs$\leftrightarrow$Uk, En$\rightarrow$Hr, Yakut$\leftrightarrow$Russian, and En$\leftrightarrow$Liv are categorized as low-resource languages.

For the WMT23 test set \cite{kocmi2023findings}, the experimental analysis covers 4 language pairs. We used the full test dataset. Among them, En$\rightarrow$De and En$\rightarrow$Ja are identified as high and medium-resource languages, with the former belonging to the same language family and the latter exhibiting significant differences. In contrast, Cs$\rightarrow$Uk and En$\rightarrow$Hr are categorized as low-resource languages, being closely related and belonging to the same language family, respectively. 

The Commonsense Reasoning MT dataset \cite{he2020box} encompasses vocabulary that requires common knowledge for resolution, along with instances of contextual/contextless grammatical ambiguity in Chinese-to-English translation data. Each translation data includes a source sentence and two contrasting translations, involving seven different types of common knowledge. Despite these elements appearing amenable to direct translation, such simplified interpretations are often misleading.


\subsection{Comparative Methods}
\label{sec:Comparative_Methods}
The following sections provide detailed descriptions of these comparisons.
\begin{itemize}[noitemsep]
    \item \textbf{Baseline}, standard zero-shot translation is performed in ChatGPT \cite{ouyang2022training} and GPT-4 \cite{achiam2023gpt} with the temperature parameter set to 0, which is the default value for our experiments.
     \item \textbf{Rerank} was conducted with the identical prompt as the baseline, employing a temperature of 0.3, in alignment with \citealp{moslem2023adaptive}. Three random samples were generated and combined with the baseline to yield four candidates. The optimal candidate was chosen through Quality Estimation (QE). 
    \item \textbf{Renfie \cite{chen2023iterative}} first requests a translation from ChatGPT, then provides the source text and translation results, and obtains a refined translation through multiple rounds of modifications by mimicking the human correction process. \textbf{Renfie\_cos} as a contrastive prompt to the \textbf{Renfie}, the work insert the word “bad” to hint that the previous translation is of low quality, regardless of its actual quality.
    \item \textbf{MAPS \cite{He2023ExploringHT}}, incorporating the knowledge of keywords, topic words, and demonstrations similar to the given source sentence to enhance the translation process, respectively.
    \item \textbf{Self-Reflect \cite{shinn2023reflexion}}, This approach requires the LLM to scrutinize and refine its translation until it deems the current output satisfactory.
    \item \textbf{MAD \cite{liang2023encouraging}} enhance the capabilities of large language models (LLMs) by encouraging divergent thinking. In this method, multiple agents engage in a debate, while a judge oversees the process to derive a final solution.
\end{itemize}

\section{Experiment Results}

\subsection{Results on Reference-free metric}
To further clarify the robustness of our evaluation, we incorporated COMET-KIWI\footnote{https://github.com/Unbabel/COMET} ~\cite{rei2022cometkiwi}, a reference-free metric in the COMET series. The experimental results are shown in Table \ref{kiwi}.
\begin{table}[!ht]\small \centering
\begin{tabular}{lcccc}
\hline
\textbf{Methods} & \textbf{En-De} & \textbf{En-Ja} & \textbf{Cs-Uk} & \textbf{En-Hr} \\ \hline
ChatGPT          &                &                &                &                \\
+Rerank          & 82.1           & 84.4           & 83.6           & 83             \\
+Self-Reflect    & 82.0           & 84.4           & 83.3           & 83.1           \\
+Dual Reflection & \textbf{82.4}  & \textbf{84.7}  & \textbf{84.2}  & \textbf{83.8}  \\ \hline
\end{tabular}
\caption{WMT22 evaluation results on COMET-KIWI metric.}
\label{kiwi}
\end{table}

These results demonstrate that our method still outperforms comparison methods in terms of COMET-KIWI scores, thereby further confirming the robustness of our evaluation. 

\subsection{Results of Additional Low-Resourced Language Pairs}
To further analyze the performance of our method in lower resource tasks, we validate the effectiveness of the DUAL-REFLECT method on 5 other lower resource languages in the WMT22 task. The experimental results are shown in Table \ref{wmt22_add_low}:
\begin{table*}[!ht]\small \centering
\begin{tabular}{lcccccccccc} 
\hline
\multirow{2}{*}{\textbf{Methods}} & \multicolumn{2}{c}{\textbf{Sah$\rightarrow$Ru}} & \multicolumn{2}{c}{\textbf{Ru$\rightarrow$Sah}} & \multicolumn{2}{c}{\textbf{Uk$\rightarrow$Cs}} & \multicolumn{2}{c}{\textbf{En$\rightarrow$Liv}} & \multicolumn{2}{c}{\textbf{Liv$\rightarrow$En}} \\ \cline{2-11} 
                                  & COMET            & BLEURT           & COMET            & BLEURT           & COMET           & BLEURT           & COMET            & BLEURT           & COMET       & BLEURT       \\ \hline
ChatGPT                           & 57.5             &  36.0                 & 52.8             &  73.2                & 88.7            & 79.0                 & 52.7             &  41.8                & 40.6        & 41.1             \\
+5shot                            & 58.3             & 36.0                 & 53.1             &  75.4                & 89.6            & 79.1                 & 55.3             & 42.1                 & 42.7        & 40.9             \\
+MAD                              & 58.1             &   37.1               & 53.5             &  76.4                & 89.6            & 79.3                 & 55.5             & 42.5                  & 43.2        &  41.3            \\
+OUR                              & \textbf{59.5}             &   \textbf{37.9}               & \textbf{54.5  }           &  \textbf{76.9}                 & \textbf{90.0}            &  \textbf{ 80.1}               &\textbf{ 56.0}             &   \textbf{43.3}               & \textbf{43.6}        &   \textbf{41.7}           \\ \hline
\end{tabular}
\caption{The main results for the WMT22 additional low-resourced language pairs are displayed. The highest values are highlighted in bold and have p-values less than 0.05.}
\label{wmt22_add_low}
\end{table*}

The experimental results demonstrate that our method improves the translation performance in terms of COMET22 and BLEURT scores for these languages, further indicating the effectiveness of DUAL-REFLECT in lower-resource translation tasks.

\subsection{Results of WMT23}
\label{sec:wmt23}
To further illustrate this point, we conducted additional experiments in WMT23 for the EN-DE , EN-JA , EN-HE, and CS-UK language pairs. The experimental results are shown in Table \ref{wmt23}:

\begin{table*}[!ht]\small \centering
\begin{tabular}{lcccccccc}
\hline
\multirow{2}{*}{\textbf{Methods}} & \multicolumn{2}{c}{\textbf{En$\rightarrow$De}} & \multicolumn{2}{c}{\textbf{En$\rightarrow$Ja}} & \multicolumn{2}{c}{\textbf{En$\rightarrow$He}} & \multicolumn{2}{c}{\textbf{Cs$\rightarrow$Uk}} \\ \cline{2-9} 
                                  & COMET           & BLEURT           & COMET           & BLEURT           & COMET           & BLEURT           & COMET           & BLEURT           \\ \hline
ChatGPT                           & 83.5            &  69.1                & 87.3            & 60.2                  & 82.1            &  69.3                & 86.7            &  74.1                \\
+5shot                            & 83.7            &  69.4                & 87.8            & 61.5                  & 82.5            &     69.8             & 87.3            & 74.5                 \\
+MAD                              & 83.9            & 70.3                 & 88.0            & 63.1                  & 82.9            & 70.0                  & 87.5            &  74.9                \\
+OUR                              & \textbf{84.3}            & \textbf{71.0 }                & \textbf{88.5 }           &   \textbf{63.6 }              & \textbf{83.1}            &   \textbf{71.7}               & \textbf{88.1}            & \textbf{75.2}                 \\ \hline
\end{tabular}
\caption{The main results from WMT23 are shown. The highest values are in bold, with p-values less than 0.05.}
\label{wmt23}
\end{table*}

Through our experiments on WMT23, we found that our method still outperforms multiple comparison methods, further demonstrating its effectiveness and generalizability.

\subsection{Human Evaluations}
\label{sec:human_evaluation}
In this section, we conduct human evaluation to measure translation quality. We assess coherence, fluency, and ambiguity resolution. Four english native speakers were invited to participate, and 50 samples were randomly selected from translations generated by different methods. For the content with Chinese ambiguity in Commonsense MT, we ensured the correctness of the source side understanding by confirming it with classmates whose native language is Chinese. For translation quality, each sentence was rated on a scale from 1 to 5, with 3 indicating a pass, 4 showing substantial consistency with the reference, and 5 being the highest score. The final score is the average of these four ratings. Additionally, in the CommonsenseMT task, the four experts scored each sample for ambiguity resolution against the reference, awarding 1 point for resolved and 0 points for unresolved.

\subsection{Case Study}
\label{sec:case_study}
\begin{figure*}[!ht]\centering
\includegraphics[scale=0.6]{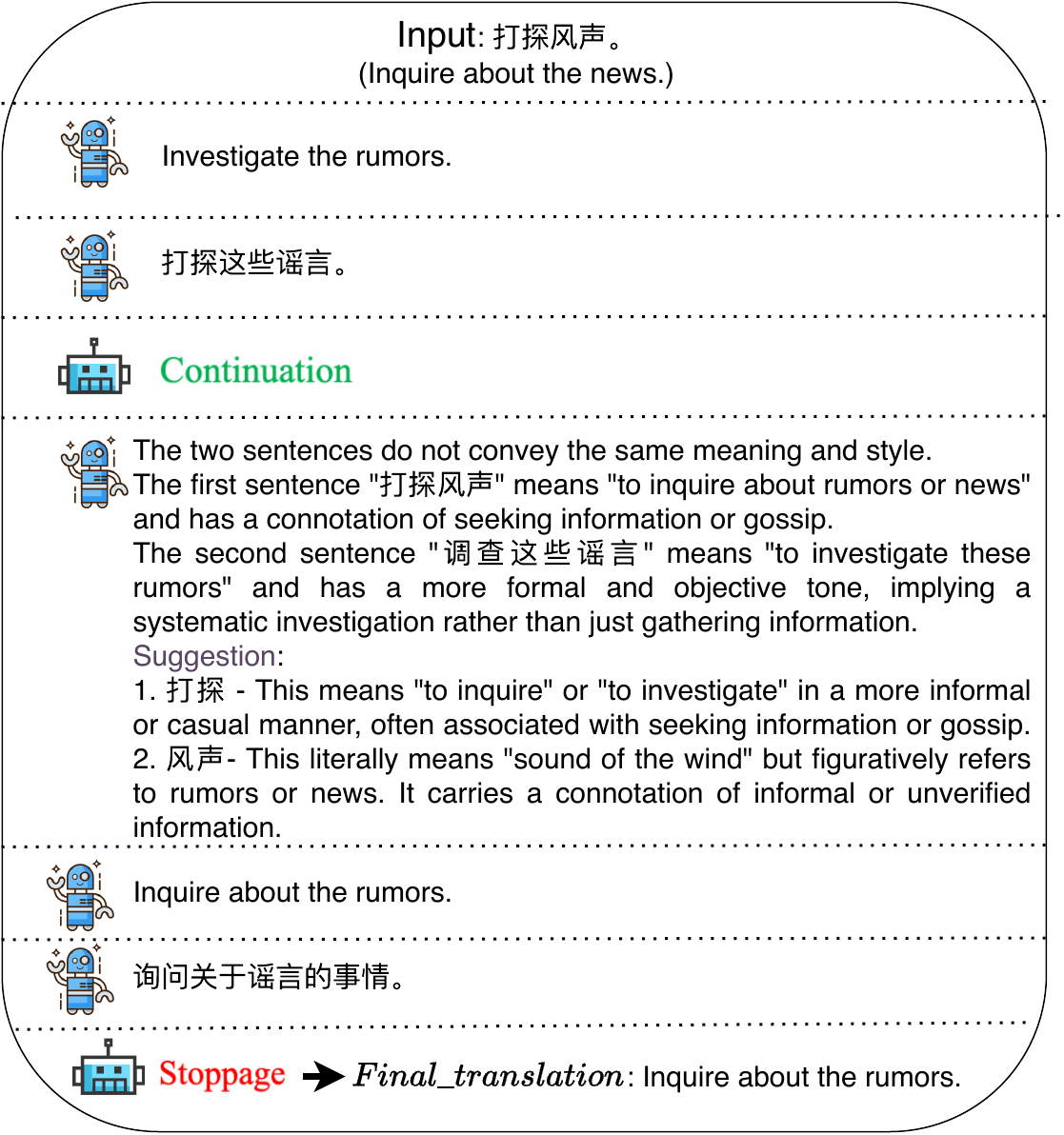} 
\caption{The DUAL-REFLECT methodology for translating \underline{positive} examples within Chinese sentences.}
\label{case1}
\end{figure*}


\begin{figure*}[!ht]\centering
\includegraphics[scale=0.6]{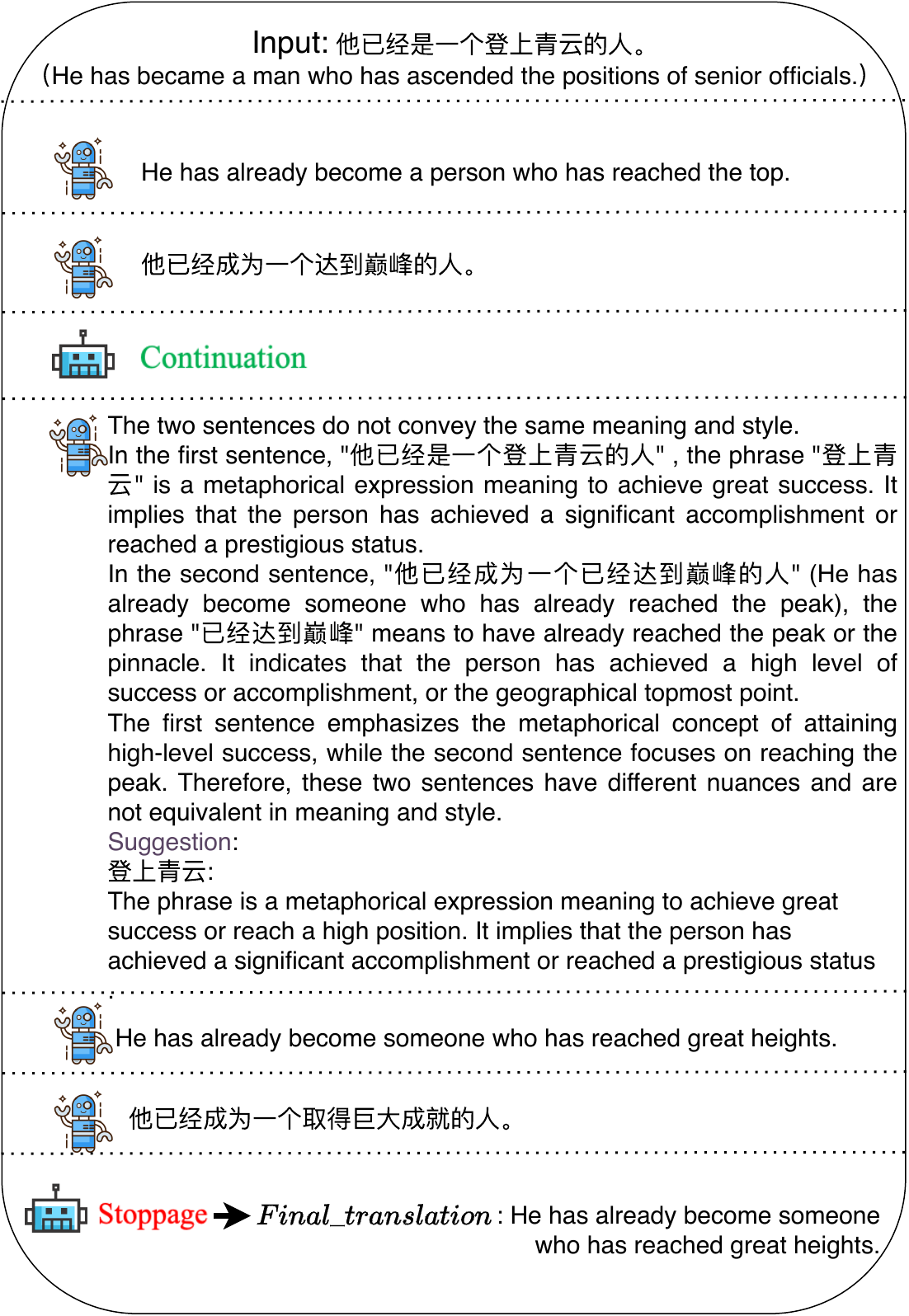} 
\caption{The DUAL-REFLECT methodology for translating \underline{positive} examples within Chinese sentences.}
\label{case3}
\end{figure*}

\begin{figure*}[!ht]\centering
\includegraphics[scale=0.6]{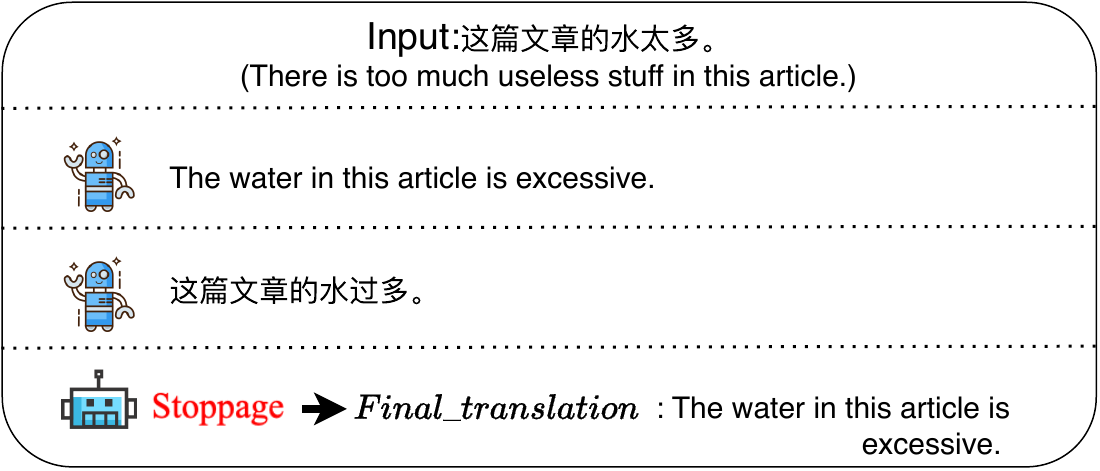} 
\caption{The DUAL-REFLECT methodology for translating \underline{negative} examples within Chinese sentences.}
\label{case4}
\end{figure*}

\begin{figure*}[!ht]\centering
\includegraphics[scale=0.6]{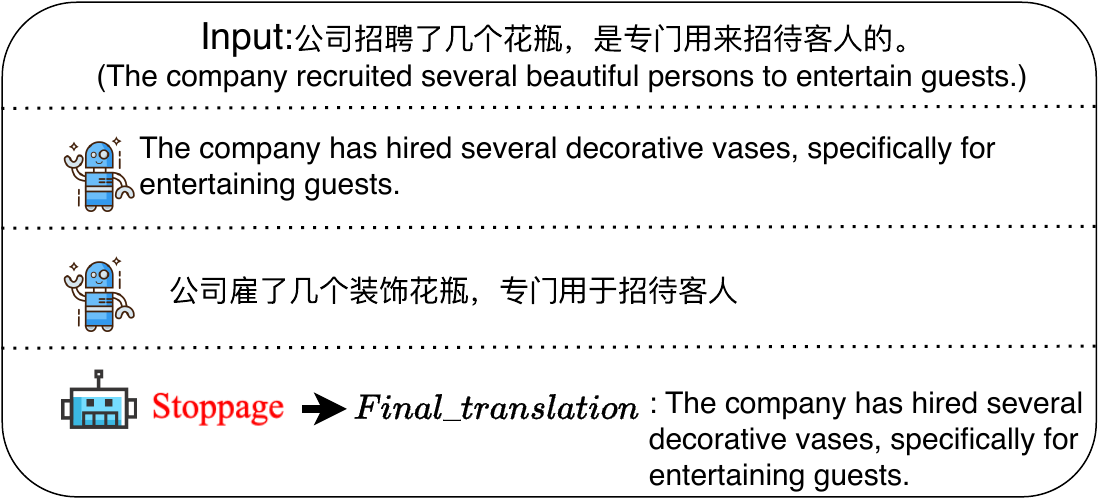} 
\caption{The DUAL-REFLECT methodology for translating \underline{negative} examples within Chinese sentences.}
\label{case5}
\end{figure*}

\end{document}